\documentclass[9pt,conference]{IEEEtran}
\IEEEoverridecommandlockouts
\usepackage{color}
\usepackage{cite}
\usepackage{amsmath,amssymb,amsfonts}
\usepackage{algorithmic}
\usepackage{graphicx}
\usepackage{subfigure}
\usepackage{textcomp}
\usepackage{xcolor}
\def\BibTeX{{\rm B\kern-.05em{\sc i\kern-.025em b}\kern-.08em
    T\kern-.1667em\lower.7ex\hbox{E}\kern-.125emX}}

\usepackage[linesnumbered, ruled, boxed]{algorithm2e}
\usepackage{framed}
 \usepackage{multirow}
\usepackage{setspace}
\usepackage{makecell}
\usepackage[misc]{ifsym}
\usepackage{fancyhdr}

 \usepackage[absolute,overlay]{textpos} 

\begin{document}
\setlength{\textfloatsep}{8pt}
\IEEEoverridecommandlockouts
\title{BlockGNN: Towards Efficient GNN Acceleration Using Block-Circulant Weight Matrices
}
\author{
{\rm Zhe Zhou\textsuperscript{1}, Bizhao Shi\textsuperscript{1}, Zhe Zhang \textsuperscript{1}, Yijin Guan \textsuperscript{1},  Guangyu Sun\textsuperscript{*}\textsuperscript{1,}\textsuperscript{2}, Guojie Luo\textsuperscript{1,}\textsuperscript{2}} \\
\textsuperscript{1}\emph{Peking University}\\
\textsuperscript{2} \emph{Advanced Institute of Information Technology, Peking University}\\
\emph{\{zhou.zhe, shi\_bizhao, zhe.zhang, guanyijin, gsun, gluo\}@pku.edu.cn}\\

}
\maketitle
\newcommand\blfootnote[1]{%
\begingroup
\renewcommand\thefootnote{}\footnote{#1}%
\addtocounter{footnote}{-1}%
\endgroup
}

\newcommand{\squishlist}{
 \begin{list}{$\bullet$}
  { \setlength{\itemsep}{0pt}
     \setlength{\parsep}{3pt}
     \setlength{\topsep}{3pt}
     \setlength{\partopsep}{0pt}
     \setlength{\leftmargin}{1.5em}
     \setlength{\labelwidth}{1em}
     \setlength{\labelsep}{0.5em} } }
\newcommand{\squishend}{
  \end{list}  }

\begin{abstract}
In recent years, Graph Neural Networks (GNNs)  appear  to be  state-of-the-art algorithms  for  analyzing  non-euclidean  graph  data. By applying deep-learning  to extract high-level representations from graph structures, GNNs achieve  extraordinary accuracy and great generalization ability
 in various tasks.  However, with the ever-increasing graph sizes,  more and more complicated GNN layers, and higher feature dimensions, the computational complexity of GNNs grows exponentially.  How to inference  GNNs in real time has become a challenging problem, especially for  some resource-limited edge-computing platforms. 

To tackle this challenge, we propose BlockGNN, a software-hardware co-design approach to realize efficient GNN acceleration. At the algorithm level, we propose to  leverage block-circulant weight matrices to greatly reduce the complexity of various GNN models.  
At the hardware design level, we 
propose a pipelined CirCore architecture,  which supports efficient block-circulant matrices computation. Basing on CirCore, we present a novel BlockGNN accelerator  to  compute various GNNs with low latency. Moreover, to determine the optimal configurations for diverse deployed tasks, we also introduce a performance and resource model that helps choose the optimal hardware parameters  automatically. 
Comprehensive experiments  on the ZC706 FPGA platform demonstrate that on various GNN tasks, BlockGNN achieves up to $8.3\times$ speedup compared to the baseline HyGCN architecture and $111.9\times$ energy reduction  compared to the Intel Xeon CPU platform.
\setlength{\skip\footins}{0.1cm}
\blfootnote{*Corresponding author.} 
\end{abstract}

\begin{textblock}{4.0}(6.2,0.50)
\Large
Accepted By DAC 2021
\end{textblock}

\section{Introduction}
\begin{figure} [t]
    \centering
    \includegraphics[width=0.98\linewidth]{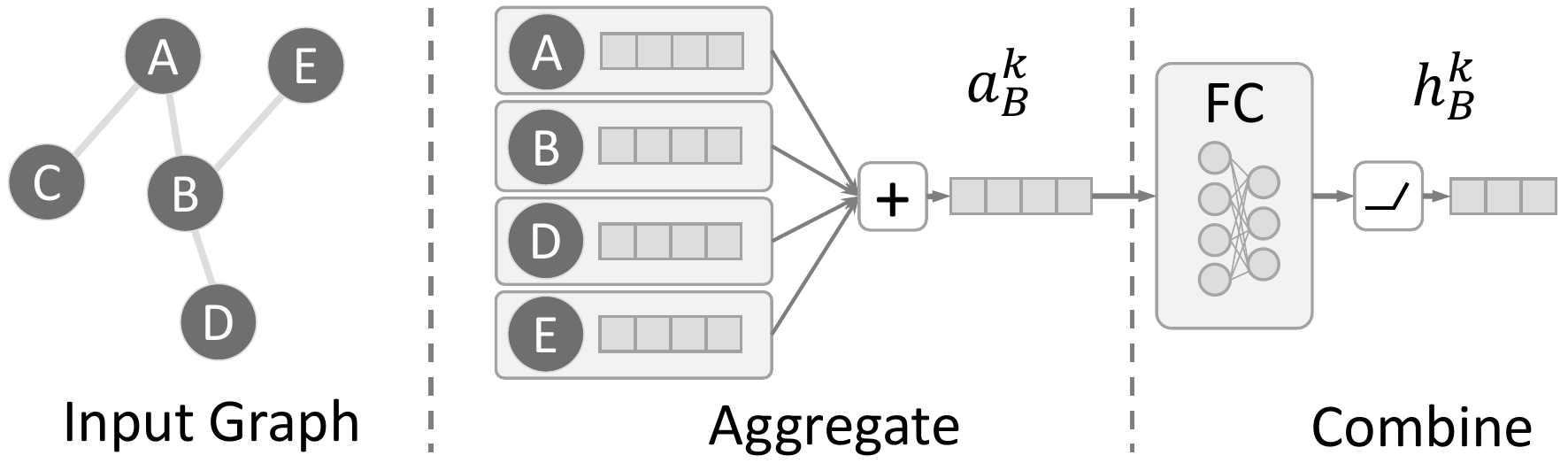} 
     \caption{The inference of GNN: a GCN example.}
                       \label{fig:gnn}
     \vspace{-4pt}
\end{figure}

Graph Neural Networks~(GNNs) have recently been proposed to apply deep learning techniques to various graph-based applications, including  node classification~\cite{Garcia-DuranN17, hamilton2017inductive}, point-cloud processing~\cite{wang2019graph}, recommendation systems~\cite{ying2018graph,fan2019graph}, IC design~\cite{wang2020gcn} and smart traffic~\cite{zhao2019t},  etc. 
Unlike traditional DNNs, GNNs are usually featured by the two-phase computing, namely \emph{Aggregation} and \emph{Combination} phases. As illustrated in Figure~\ref{fig:gnn}, to compute the hidden feature vector  of node $B$ in the $k^{th}$ layer with a Graph Convolutional Network (GCN)~\cite{kipf2016semi}, the \emph{Aggregation}  phase  collects feature vectors from the neighbors of node $B$ and sums them up, while the \emph{Combination} phase updates the aggregated  feature using a fully-connected layer. By stacking $K$ such layers, a GNN model can extract $K$-hop information from the input graph and finally produces node-level representations for various down-stream tasks. 

Besides GCN, many other GNN algorithms are also proposed, such as G-GCN\cite{2017Encoding}, GraphSAGE~\cite{hamilton2017inductive}, and Graph Attention Network (GAT)\cite{velivckovic2017graph}. These GNNs develop towards higher accuracy and better generalization ability. However, the  ever-increasing graph sizes~\cite{chiang2019cluster}, more and more complicated GNN architectures~\cite{hamilton2017inductive,velivckovic2017graph,2017Encoding}, and higher dimensionality of hidden features also make it challenging to inference these GNNs with low latency, especially for some resource-limited edge-computing platforms. For instance, in some smart-traffic scenarios, the deployed edge servers need to predict traffic timely using GNNs~\cite{zhao2019t}. Smart vehicles also leverage GNNs to detect 3D objects from LiDAR point cloud data in real time~\cite{Shi_2020_CVPR}. 
Therefore, it is urging to develop efficient GNN acceleration approaches.   
   
Many domain-specific architectures have been proposed to partially address the challenges in GNN computing. For instance, AWB-GCN~\cite{geng2019awb} and HyGCN~\cite{yan2020hygcn} are two accelerators that mainly target GCN acceleration.  EnGN~\cite{engn} is another work which is designed to accelerate various GNN algorithms with a three-stage architecture.
However, we notice that these previous works only focus on hardware-level and system-level designs to improve the processing efficiency, while the GNNs still have tremendous computational complexity, making real-time inference hard to achieve. For instance, the popular GraphSAGE algorithm with the mean-pooling aggregator (GS-Pool)~\cite{hamilton2017inductive} requires about 1.9 trillion floating-point operations per-layer when used on \emph{Reddit} dataset, which is still unbearable for many resource-limited edge-computing platforms. What's worse, the deployed tasks usually have different GNN models and input feature dimensions. Using fixed hardware configurations may result in sub-optimal performance on the changeable tasks. 
 
To tackle these challenges, in this paper we propose BlockGNN, a software-hardware co-design approach to accelerate various GNNs. Specifically, at the algorithm level, we apply block-circulant weight matrices to compress  GNN models. By imposing the block-circulant constraint on weight matrices during training and leveraging Fast Fourier Transform (FFT) to accelerate the inference, we reduce the computational complexity from $O(n^2)$ to $O(nlogn)$ with negligible accuracy drop.  
At the hardware design level, we propose a pipelined CirCore architecture to accelerate the computation of block-circulant matrices efficiently. 
Basing on CirCore, we present a novel BlockGNN accelerator to compute various GNNs with low-latency. Moreover, to determine the optimal configurations for diverse deployed tasks, we also introduce a performance and resource model that helps choose the optimal hardware parameters automatically. Comprehensive experiments using the Xilinx ZC706 FPGA platform demonstrate that on various GNN tasks,  BlockGNN achieves up to $8.3\times$ speedup  and $111.9\times$ energy reduction compared to the baseline HyGCN architecture and Intel Xeon 5220 CPU, respectively.

To summarize, this article makes the following key contributions:
\begin{itemize}
    \item We propose to leverage block-circulant weight matrices to compress various GNNs. Extensive algorithm-level experiments demonstrate the efficiency of this approach.
    \item We design a pipelined CirCore architecture to efficiently compute block-circulant matrices. Based  on  CirCore, BlockGNN accelerator is presented to process  GNNs with low latency. 
    \item We introduce a performance and resource model to determine the optimal hardware parameters  automatically according to deployed GNN tasks.
    \item We implement a prototype of our design  on ZC706 FPGA and demonstrate its superiority against several baselines.
\end{itemize}

\section{Preliminaries and Motivations}



\subsection{Graph Neural Networks} GNNs are now widely adopted to learn high-level representations from graphs with deep-learning techniques~\cite{wu2020comprehensive}. Generally, a GNN is composed of several layers, each layer has two computational phases: 1) neighbor aggregation phase and 2) feature combination phase. Figure \ref{fig:gnn} has illustrated this procedure with a simple GCN example.  We  also formulate the inference of each GNN layer as follows:
\begin{align}
    \label{eq:aggregate}
    a_v^{k}&=\textsc{Aggregate}(h_u^{k-1}|u\in \mathcal{N}v)\\
    \label{eq:combine}
    h_v^{k}&=\textsc{Combine}(h_v^{k-1},a_v^{k})
\end{align}

\newcommand{\tabincell}[2]{\begin{tabular}{@{}#1@{}}#2\end{tabular}}

\begin{table}[]
\setlength{\abovecaptionskip}{0cm} 
		\setlength{\belowcaptionskip}{-0.2cm}
\caption{GNN Algorithms}
\label{tab:gnn}
\resizebox{0.485\textwidth}{!}{
\renewcommand{\arraystretch}{1.65}
\begin{tabular}{c|c|c}
\hline
Variant & Aggregation & Combination \\ \hline
GCN~\cite{kipf2016semi}       &$a_v^{k}=\sum_{u\in \mathcal{N}_v}h_u^{k-1}*V_{degree}^{-1/2}$&           $Relu(\mathbf{W}^{k}a_{v}^{k})$            \\ \hline
GS-Pool~\cite{hamilton2017inductive}  &     $a_v^{k}=max_{u\in \mathcal{N}_u} Relu(\mathbf{W}_{pool}^{k}h_{u}^{k}+b^{k})$               &      $Relu(\mathbf{W}^{k}(a_{v}^{k}|h_v^{k}))$                 \\ \hline
G-GCN~\cite{2017Encoding}   &     
\tabincell{c}{$ \eta_{u}=\sigma(\mathbf{W}_{H}^{k}h_{u}^{k-1}+\mathbf{W}_C^{k}h_v^{k-1})$ \\ $a_v^k = \sum_{u\in\mathcal{N}_v}\eta_u \odot h_u^{k-1}$ }
              &      $Relu(\mathbf{W}^{k}a_{v}^k)$                 \\ \hline
GAT~\cite{velivckovic2017graph}       & 
\tabincell{c}{$\alpha^k_{ij}=softmax(a(\mathbf{W}h^{k-1}_i,\mathbf{W}h^{k-1}_j))$\\$a_v^k=\sum_{j\in N_i}\alpha_{ij}h^{k-1}_j$ }
              &       $Elu(\mathbf{W^k}a^{k}_v)$               \\ \hline
\end{tabular}}
\end{table}

\noindent In the formulations, $h_v^{k-1}$ denotes the hidden feature of node (vertex) $v$ at the $k$-$1^{th}$ layer ($k=1,2,...,K$). Thus the input feature of $v$ is $h_v^{0}$.  The neighbors of node $v$ in the graph is denoted as $\mathcal{N}_{v}$. For each node $v$, \textsc{Aggregate}  (Equation \ref{eq:aggregate}) gathers features from the neighbor nodes using \emph{sum}, \emph{average}, or other  functions in Table \ref{tab:gnn}. $a_v^{k}$ is the  aggregation results at the $k^{th}$ layer, which  is then fed into a combiner (usually a fully-connected layer)  to get the transformed feature $h_v^{k}$. After finishing the computation of the $k^{th}$ layer, $h_v^{k}$ further serves as the input of the next layer. By stacking $K$ such layers, GNN models can extract $K$-hop information from given graph structures and finally produce node-level representations for down-stream tasks.

To pursue higher accuracy and better generalization ability, various GNN algorithms have  been proposed, 
Table \ref{tab:gnn} listing some popular ones. As we can see, current GNN algorithms adopt diverse aggregation and combination functions. 
We can also conclude from the table that the combination phases are mainly matrix-vector multiplications. While for aggregators, the four algorithms differ from each other.  In the GCN model, the aggregation phase only involves vector-scalar multiplication and element-wise addition. On the contrary, the rest of algorithms all have one or more matrix-vector multiplications in their aggregators.  By adopting learnable weight matrices in the aggregation phases, these more complicated GNNs tend to achieve higher accuracy but demand much more computations, challenging many resource-limited edge-computing platforms.  To quantitatively analyze the potential bottlenecks of running these GNN models, we  conduct profiling  in the following subsection.

\subsection{The Profiling of GNNs}
\label{sec:profiling}

Considering that the computational overhead
and bandwidth requirements are the main concerns for real-time
applications,  we evaluate the GNN algorithms with respect to the total computations and arithmetic intensity. Specifically, we choose the widely-used \emph{Reddit} dataset containing $233$K nodes  and $115$M edges to conduct the profiling.  We adopt the  sampling-based aggregation strategy~\cite{hamilton2017inductive} for all algorithms, where the sample size is $25$. We set both the input feature and output feature to $512$ dimensions, which are typical values in GraphSAGE. For GAT, we assume two $128$-dimensional attention heads for evaluation. 
Given that \emph{Aggregation} and \emph{Combination} phases are usually considered to have different execution patterns~\cite{yan2020hygcn}, they are evaluated individually. We analyze the total computations (FLOPs) and arithmetic intensity (Operations per Byte) of the four algorithms. The results are shown  in Table~\ref{tab:gnn_profile}. We find that  the  character of GCN is consistent with the previous work~\cite{yan2020hygcn} in that the aggregation phase has a low arithmetic intensity while the combination phase is computation-intensive. However, unlike GCN, the other three GNN models all involve tremendous computations in both the aggregation and combination phases, which is due to the adoption of  weight matrices in their aggregators. Upon stacking several layers together or taking larger graphs as input, the total computation will easily exceed many resource-limited platforms' compute capacity. The problem will get even worse when real-time inference is the main concern. Therefore, it is vital to take some practical approaches to accelerate the GNN inference.


\begin{table}[]
\setlength{\abovecaptionskip}{0cm}
		\setlength{\belowcaptionskip}{-0.2cm}
\caption{GNN Profiling}
\label{tab:gnn_profile}
\centering
\resizebox{0.48\textwidth}{!}{
\renewcommand{\arraystretch}{1.1}
\begin{tabular}{|c|c|c|c|c|}
\hline
\multirow{2}{*}{Algorithm} & \multicolumn{2}{c|}{Total Computations} & \multicolumn{2}{c|}{Arithmetic Intensity} \\ \cline{2-5} 
                           & Aggregation     & Combination    & Aggregation            & Combination            \\ \hline
GCN                        & $3.7\times10^{9}$         & $7.5\times10^{10}$      &                     $0.5$   &        $256.3$            \\
GS-Pool                    & $\mathbf{1.9\times10^{12}}$       & $1.5\times10^{11}$       &    $257.5$                  &        $512.2$           \\
G-GCN                  &  $\mathbf{3.7\times10^{12}}$       & $7.5\times10^{10}$       &    $256.0$                  &       $256.3$             \\
GAT                        & $\mathbf{1.9\times10^{12}}$        &  $7.5\times10^{10}$          &  $512.8$                 &    $256.3$                \\ \hline
\end{tabular}
}
\end{table}

\subsection{GNN Acceleration}

Recently, several domain-specific architectures have been proposed to partially address the challenges in GNN inference. To name a few, AWB-GCN~\cite{geng2019awb} views GCN inference as matrix-multiplications and proposes an auto-tuning workload balancing mechanism to address the issue of workload imbalance.  HyGCN~\cite{yan2020hygcn} is another work that mainly focuses on GCN algorithms. Unlike AWB-GCN, it computes GCNs in the spatial domain and proposes a two-stage  accelerator to deal with both the memory-intensive aggregation phase and compute-intensive combination phase.  EnGN~\cite{engn}, on the contrary,  is a general-purpose accelerator designed for  various GNN algorithms with a specialized three-stage architecture. 

However, all these works do not explicitly reduce the computational complexity of GNNs. That is to say, when processing large graphs with complicated GNN models, the compute capacity will easily become the bottleneck, according to our analysis in Section~\ref{sec:profiling}. This problem will become even worse when deploying GNN tasks on some resource-limited  edge-computing platforms, where  the real-time performance, energy-efficiency and hardware resources budget are equally important. To tackle this challenge, we propose a software-hardware  co-design  approach to efficiently compress various GNN models and realize cost-efficient inference through a novel hardware architecture. Both  algorithm  details and the hardware design will be introduced in the next section.

\section{Proposed Approach}
In this section, we propose BlockGNN, a software-hardware co-design approach to achieve efficient GNN acceleration. Firstly, we introduce how to use block-circulant weight matrices to compress various GNNs. Secondly, we propose the BlockGNN architecture to efficiently compute the compressed GNNs. Finally, a performance and resource  model is  introduced to  help determine the optimal hardware parameters  for different GNN tasks automatically.  

\subsection{Block-Circulant Weight Matrices for GNNs}
\begin{figure} [t]
\setlength{\abovecaptionskip}{-0.1cm} 
    \centering
    \includegraphics[width=0.99\linewidth]{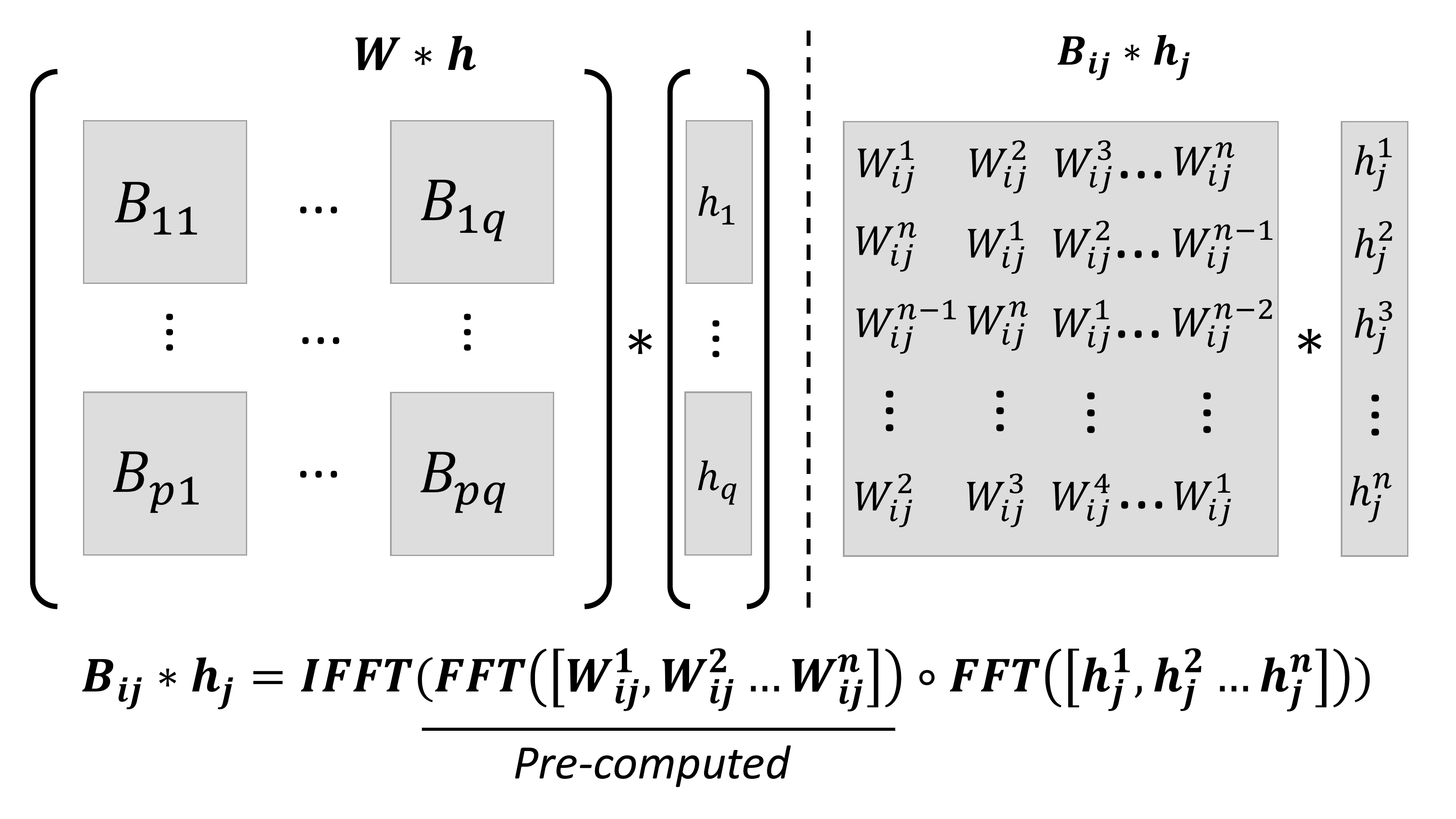} 
      \caption{Block-Circulant Weight Matrices Compression. The original weight matrix $W$ with $N$ rows and $M$ columns is divided to $p*q$ circulant blocks. The feature vector $h$ is also divided to $q$ sub-vectors.}
            \label{fig:bcm}
\end{figure}

From Table~\ref{tab:gnn} and  Table~\ref{tab:gnn_profile} we can find that the  multiplications of weight matrices and feature vectors dominate the GNN computing. Thus, 
how to reduce the computational complexity of these matrix-vector multiplications is the first  concern. For effective GNN compression,  here we propose to leverage block-circulant weight matrices, a structured compression method designed for traditional DNNs~\cite{CirCNN,PSB,wang2018c}. To the best of our knowledge, we are the first to study its usage in the graph neural network domain.

First of all, we illustrate the computation of block-circulant matrices in Figure~\ref{fig:bcm}. Consider the multiplication of weight matrix $W$ and feature vector $h$ in a GNN layer. The original computational complexity is $O(NM)$, where $N$ and $M$ represent the output and input  dimensions (rows and columns of matrix $W$).  If $W$ is  a block-circulant matrix, it is partitioned into several $n\times n$ blocks, $p\times q$ in total, where  $p =
\frac{N}{n}$, $q = \frac{M}{n}$ (Note, if $M$ or $N$ is not divisible by $n$, just use zero-padding). Each block has the so-called \emph{circulant structure}. That is to say, in any block $B_{ij}$, $i\in\{1...p\}$, $j\in\{1...q\}$, all the other rows are circulant permutations of the first row. This block-circulant property is guaranteed by adding certain constraints during model training. 
The benefit of adopting block-circulant matrices is that we can use FFT to accelerate the calculation~\cite{CirCNN}. As shown in the formulation of Figure~\ref{fig:bcm}, for the computation of each block, we first transform the block into the spectral domain by applying FFT to the first row, so does the corresponding feature vector. Then the spectral weight and feature are multiplied in an element-wise manner. We transform the multiplication results back to the spatial domain through Inverse FFT (IFFT) for further inter-block accumulation and finally produce the output vector.  By doing so, the original $O(n^2)$ matrix-multiplication is replaced by $O(nlogn)$ FFT operations. The storage complexity of the weight matrices is also reduced from $O(n^2)$ to $O(n)$, because we only need to store the first row of each block. 

Actually, before inference, the trained $W$ can be transformed into the spectral domain in advance (We use $\widehat{W}$ to denote the pre-computed spectral weight), thus only $\emph{FFT}([h_j^1,h_j^2...h_j^n])$ has to be  calculated on-the-fly.
We also notice that since IFFT is a linear transformation, then we have 
$\sum_{i=1}^N \emph{IFFT}(h_i') =  \emph{IFFT}(\sum_{i=1}^N h_i')$.
This equation indicates that the original compute flow in \cite{CirCNN} that conducts inter-block  accumulation in the spatial domain can be further simplified. By accumulating directly in the spectral domain,  the amount of IFFT operations is reduced from $p\times q$ to $p$. This point is also noticed by a previous work~\cite{wang2018c}. 
Having the above observations, we finally derive the optimized  procedure  in Algorithm~\ref{algo:1}. 

\subsection{Compression Ratio and Accuracy}
\begin{table}[]
\centering
\setlength{\abovecaptionskip}{0cm}
		\setlength{\belowcaptionskip}{-0.2cm}
\caption{Algorithm Evaluation Results}
\label{tab:compression_results}
\small
\resizebox{0.49\textwidth}{!}{
\begin{tabular}{|l|rr|cccc|}
\hline

Block Size & TCR &SR~~ & GCN & GS-Pool & G-GCN & GAT \\
\hline

$~\ n=1$ & $1.0\times$ & $1.0\times$ & 0.924 & 0.948 & 0.950 & 0.926 \\
$~\ n=16$ & $4.0\times$ & $16.0\times$ & 0.922 & 0.941 & 0.944 & 0.922 \\
$~\ n=32$ & $6.4\times$ & $32.0\times$ & 0.920 & 0.939 & 0.942 & 0.921 \\
$~\ n=64$ & $10.7\times$ & $64.0\times$ & 0.920 & 0.938 & 0.938 & 0.919\\
$~\ n=128$ & $18.3\times$ & $128.0\times$ & 0.919 & 0.938 & 0.935 & 0.920 \\
\hline

\end{tabular}

}
\vspace{1pt}
~\\
 TCR: Theoretical Computation Reduction. SR: Storage Reduction.
 \vspace{-0.5em}
\end{table}

\begin{algorithm}[t]
\small
\label{algo:1}
\caption{Block-Circulant Matrix Computation}
\textbf{Input:} $\widehat{W},h,p,q,n$,\\
\textbf{Output:} $h'$\\
\For{i~$\leftarrow$~\emph{1} until $p$}{
Initialize $h_i'$ with zeros.\\ 
\For{j~$\leftarrow$~\emph{1} until $q$}{
    $h'_i = h'_i+  [{\widehat{W}_{ij}^{1}},...,{\widehat{W}_{ij}^{n}}]\circ \emph{FFT}([h_j^1,...,h_j^n])$ 
    }
    $h_i'=\emph{IFFT}(h'_i)$
}
\textbf{Return} $h'$
\end{algorithm}

The block-circulant weight matrices are proved to have the  \emph{universal approximation property}~\cite{CirCNN}, which means it can solve a broad range of  machine learning problems. However, we are still curious about the  compression ratio and   accuracy when applied to GNN tasks. Therefore, we evaluate  block-circulant matrices on various GNN algorithms. We implement the GNN algorithms as well as the block-circulant compression logic basing on GraphSAGE, a popular mini-batch based  training framework~\cite{hamilton2017inductive}. We evaluate the accuracy using \emph{Reddit} node-classification dataset. Each GNN model has two layers, and the hidden-vectors' dimensions are set to $512$. To study the model accuracy under different compression ratios, we choose several block sizes $n$, ranging from $16$ to $128$. For consistency, we use $n=1$ to denote the original uncompressed models. 

We list the storage reduction, theoretical computation reduction, and attainable accuracy in Table~\ref{tab:compression_results}.  A larger block size $n$  brings a higher compression ratio but usually suffers from a more massive accuracy drop. However, even when $n$ is set to $128$, the accuracy drop is within $1.5\%$. Such trade-off is beneficial  for many real-time scenarios, where the speedup is more critical  than the negligible accuracy drop. In conclusion, we have proved that block-circulant matrices is a practical compression method for various GNN algorithms.





\subsection{BlockGNN Architecture}

\label{Architecture}

\begin{figure} [t]
    \setlength{\abovecaptionskip}{-0.1cm} 

    \centering
    \includegraphics[width=0.9\linewidth]{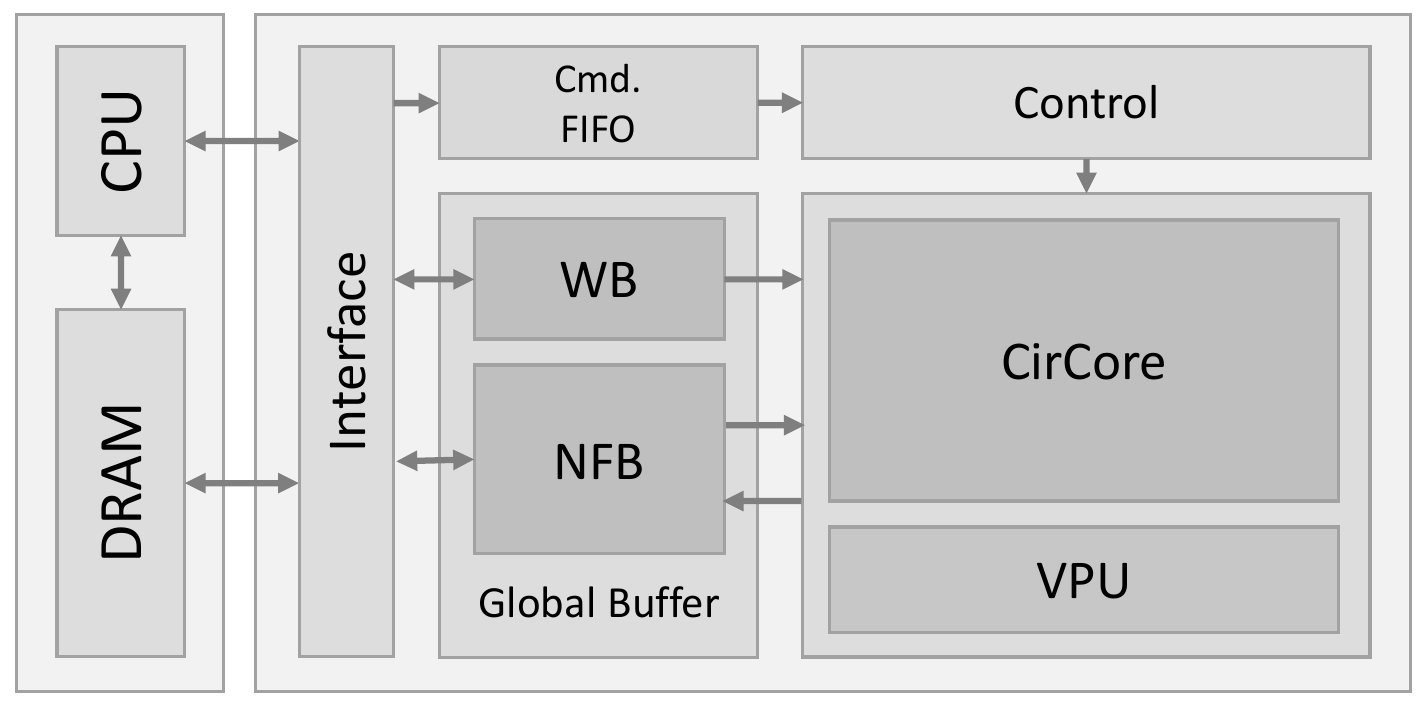} 
      \caption{BlockGNN System.}
            \label{arc}
            \vspace{-0.5em}
\end{figure}

\begin{figure} [t]
\label{bcm_computation}
\setlength{\abovecaptionskip}{-0.1cm} 
    \centering

    \includegraphics[width=0.9\linewidth]{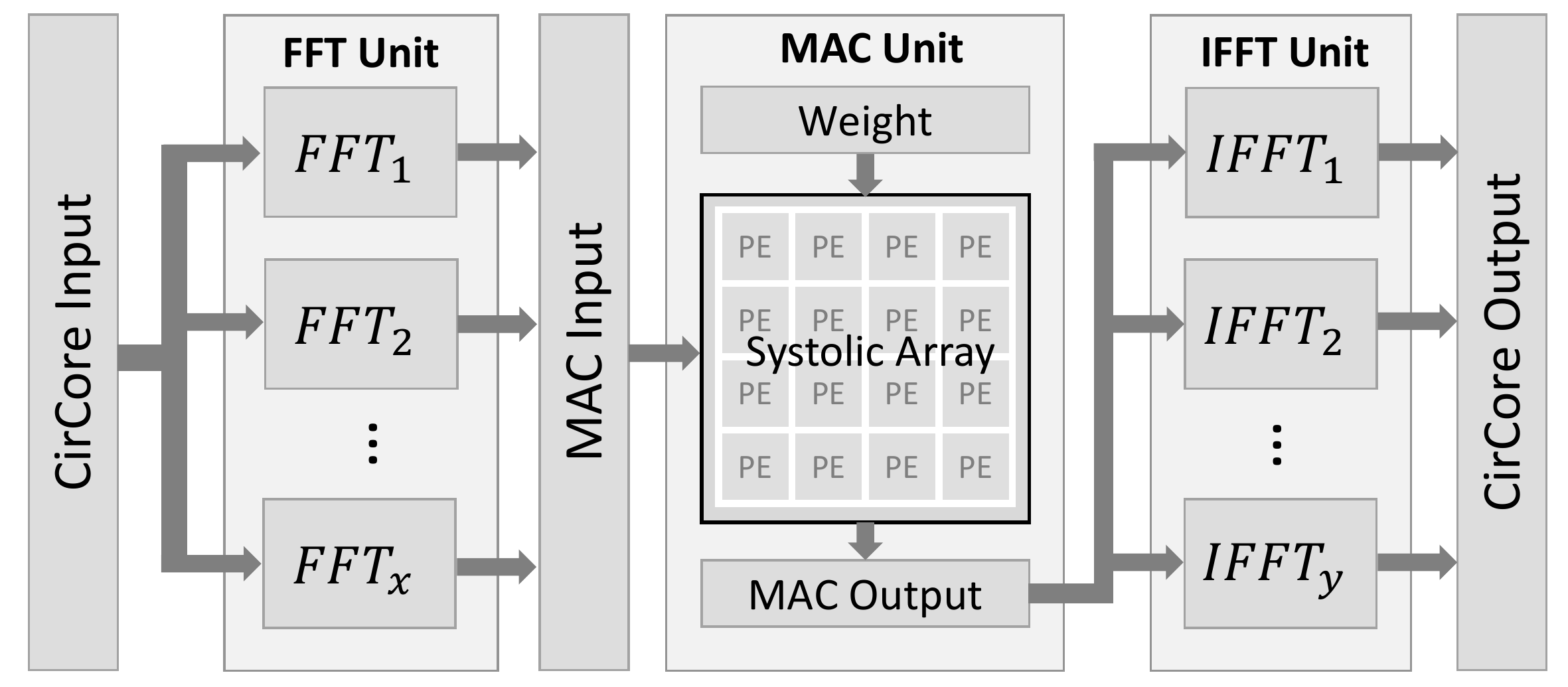} 
      \caption{CirCore Architecture.}
            \label{compute_core}
                   \vspace{-0.5em}
\end{figure}

\begin{figure} [t]
    \centering
        \setlength{\abovecaptionskip}{-0.08cm} 
    \includegraphics[width=0.9\linewidth]{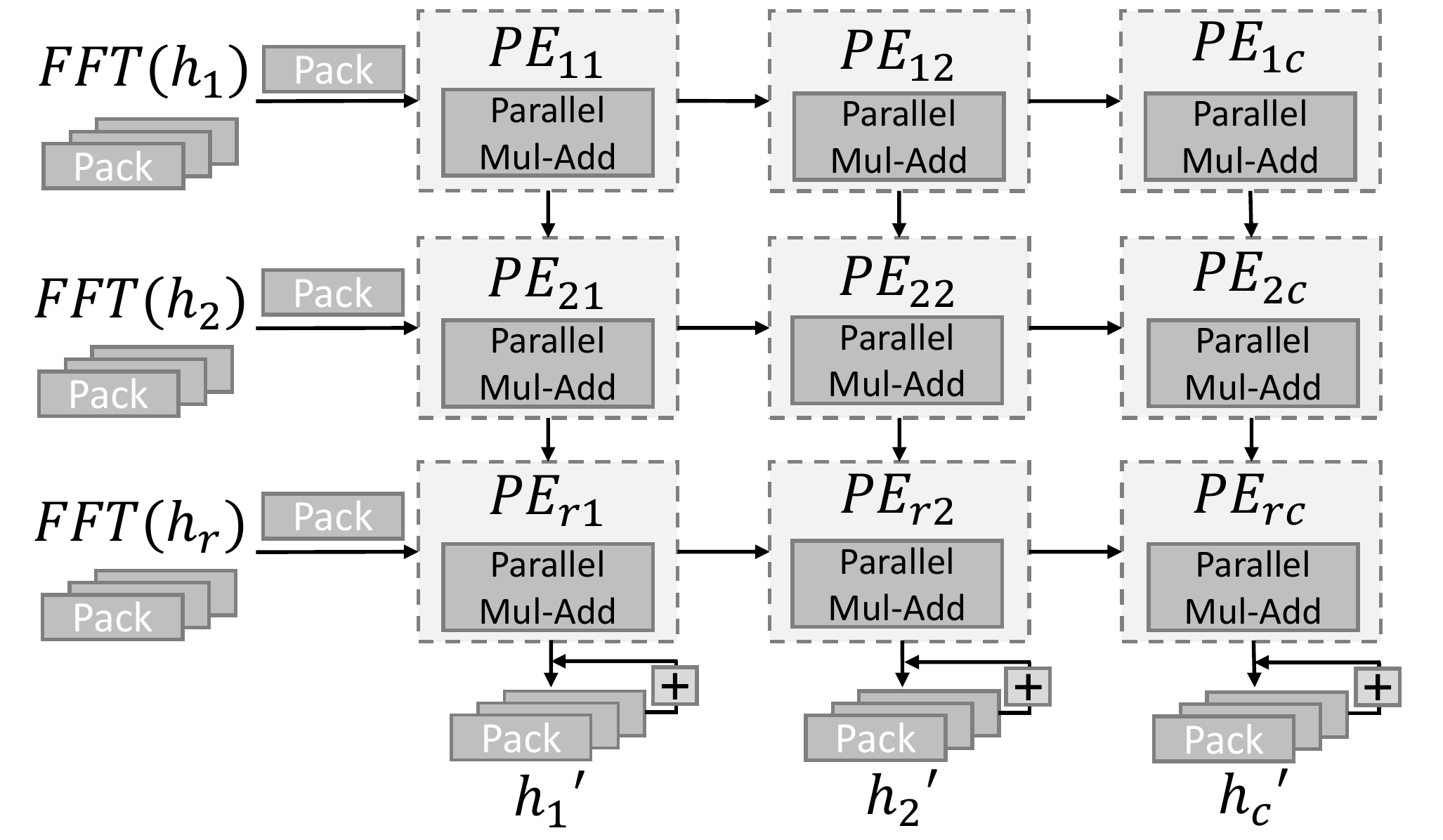} 
      \caption{Systolic Array Dataflow.}
            \label{systolic}
                \vspace{-0.5em}
\end{figure}

In the previous subsection we have proved the effectiveness of applying block-circulant compression to GNN tasks. However, another problem emerges: how to process the compressed GNN models as efficiently as possible? To answer this question, we propose a novel BlockGNN accelerator.  The overall system is shown in Figure \ref{arc}. 
BlockGNN adopts the widely-used vertex-centric processing paradigm. In each compute pass, the host CPU samples a batch of neighbor nodes and sends the corresponding features to the BlockGNN accelerator, as well as the control commands. The accelerator side conducts aggregation and combination according to the received commands and sends the updated node features back to the host side DRAM. 
To implement such workflow efficiently, we design several key components in the BlockGNN accelerator, including \textbf{CirCore}, \textbf{VPU (Vector Processing Unit)}, and \textbf{Global Buffer}, etc. In the rest of this section, we will introduce these components separately.

$\bullet$~\noindent\textbf{CirCore:}
To compute  block-circulant weight matrices with high throughput, we propose a pipelined CirCore architecture. As shown in Figure \ref{compute_core}, CirCore is organized as a three-stage pipeline. The first stage is an FFT unit that transforms feature vectors from the spatial domain to the spectral domain. To improve the throughput, it consists of $x$  FFT channels to  compute $x$ sub-vectors in parallel. Note, instead of assigning different feature vectors to each channel, we explore intra-vector parallelism first. The sub-vectors $\{h_1,h_2,...h_q\}$ of the same vector $h$ are dispatched to different channels simultaneously. Only if $x$ is greater than $q$, sub-vectors from another $h$ are dispatched to the remaining channels. Thus, to guarantee high utilization,  $x$ is better to be an integral multiple of $q$.

According to Algorithm~\ref{algo:1}, after transforming sub-vectors $\{h_1,h_2,...h_q\}$ to the frequency domain, we need to perform element-wise multiplication and accumulation with the weight vectors $\widehat{W}_{ij}$, $i\in\{1...p\}$, $j\in\{1...q\}$.
We design a systolic array architecture as the second stage to perform such multiply-accumulate (MAC) operations.
Figure~\ref{systolic} illustrates the designed systolic array dataflow. As depicted, there is an $r\times c$ PE array, which takes $r$ spectral sub-vectors as input and produces $c$ sub-vectors as output. The spectral sub-vectors are further partitioned into smaller packs, each containing $l$ elements. The pack size is equal to the port width of each PE in the systolic array.  For each layer, the inputs are multiplied with the same matrix, so the systolic array adopts a \emph{weight-stationary} dataflow. 
The weight vectors $\widehat{W}_{ij}$ are pre-loaded to the corresponding PEs. Upon starting the computation, 
the input packs flow horizontally while the accumulated values (also packed) flow vertically.  In each PE, there is a \emph{Parallel Mul-Add} unit to compute $l$ element-wise multiplication and accumulation in parallel. Therefore, we are able to adjust the computational capacity  through setting the parallel parameter $r, c$ and $l$.  By this mean, if the  systolic array is fully-loaded, it is able to  process a feature vector every  $ \lceil\frac{q}{r}\rceil \times \lceil \frac{p}{c} \rceil \times \lceil \frac{n}{l} \rceil $ cycles. 

The last stage of the pipeline is an IFFT unit with  $y$ parallel channels to transform the results back  to the spatial domain. Each channel's IFFT core has the same architecture as the  FFT core but loads  different twiddle factors. Finally, to balance this three-stage pipeline, we need to carefully determine the hardware resources allocated to each stage. We propose a performance and resource model to choose the hardware  parameters $x,y,r,c$ and $l$ of CirCore automatically. We will discuss this in the next subsection.

$\bullet$~\noindent\textbf{VPU:}
We set a VPU to support  non-linear functions (eg. \emph{ReLU}, \emph{Exp} and  \emph{Sigmoid}) and other vector-level computations (vector-vector multiplication and vector-vector addition). VPU is necessary for the aggregation operations of some GNN algorithms such as GCN. Moreover, VPU  takes the responsibility of adding  bias to the outputs.  Essentially,  the VPU is designed as a SIMD unit. Similar with \cite{yan2020hygcn}, we assume the VPU has $m$  lanes, each processing 16 real-valued  elements in parallel. Given a GNN task, the optimal $m$ is also determined by our performance model.

$\bullet$~\noindent\textbf{Global Buffer:}
We set a global buffer to hold both the weight and node features. As shown in Figure~\ref{arc}, the global buffer is further partitioned into a \emph{Weight-Buffer} (WB) and a \emph{Node-Feature-Buffer}~(NFB). The WB stores spectral weight matrices $\widehat{W}$ of each layer. With the block-circulant compression, we only need to store $\frac{1}{n}$ of  original weights. We use NFB to buffer the input and updated features.  We load data in and off NFB in batch  and use ping-pong buffers to hide the data-loading latency. 
 Note that although HyGCN \cite{yan2020hygcn} equips a large eDRAM cache to improve data reuse, our profiling reveals that for running heavy GNNs  on resource-limited edge platforms, computation is the primary bottleneck. Therefore, we just leverage node prefetching  to fully utilize the memory bandwidth rather than adopting complicated caching mechanism. 




\subsection{Performance and Resource Model}
\label{performance_model}
We notice that the GNN models, hidden feature dimensions, and input graphs vary significantly  in different deployed tasks. Thus, using a single set of hardware parameters can not always guarantee optimal performance. Given a GNN model and input graph, we propose a performance and resource model to help determine several hardware parameters, including FFT/IFFT channels $x, y$, systolic array shape $r, c$, PEs' parallel degree $l$,  and total  VPU lanes $m$. It is especially useful for FPGA-based re-configurable implementations. 

Take GS-Pool model as an example and suppose an input graph $G$ with $|\mathcal{V}|$ nodes and  $K$ layers (typically, $K=2$). In the aggregation phase of the $k^{th}$ layer ($k= 1,...,K$),  
we sample $ S_{(k)}$ neighbours for each target node. 
The original weight matrix is  $N_{(k)}\times M_{(k)}$, and the block size is $n$, so we have $p_{(k)}=\frac{N_{(k)}}{n},q_{(k)}=\frac{M_{(k)}}{n}$.
We assume the required cycles to process $n$ elements per FFT channel is $\alpha(n)$, which is determined by the FFT core's hardware implementation. Then we can use a formula to estimate the total FFT cycles: 
\begin{equation}
\setlength{\abovedisplayskip}{4pt}
\setlength{\belowdisplayskip}{4pt}
    cycle\_\mathit{fft}_{(k)} =  \alpha(n) \times \lceil\frac{  S_{(k)} \times  q_{(k)}   }{x}\rceil 
\end{equation}

\noindent According to our design, the  systolic array has $r\times c$ PEs, each processing $l$ element-wise MAC in parallel. Therefore, if the systolic array is  fully loaded, we can estimate the cycles:

\begin{equation}
\setlength{\abovedisplayskip}{3pt}
\setlength{\belowdisplayskip}{3pt}
    cycle\_{mac}_{(k)}=  S_{(k)} \times  \lceil \frac{q_{(k)}}{r} \rceil \times \lceil \frac{p_{(k)}}{c} \rceil \times  \lceil \frac{ n  }{l} \rceil
\end{equation}

\noindent  Since IFFT and FFT have the same hardware logic,  the overhead to process $n$ elements per IFFT channel is also $\alpha(n)$. Thus we can estimate the total IFFT cycles: 

\begin{equation}
\setlength{\abovedisplayskip}{3pt}
\setlength{\belowdisplayskip}{3pt}
    cycle\_\mathit{ifft}_{(k)} =  \alpha(n) \times \lceil\frac{  S_{(k)} \times  p_{(k)}   }{y}\rceil 
\end{equation}


\noindent The VPU has $m$ lanes, each containing a SIMD-16 unit. We assume the cycles needed to calculate max-pooling among $S_{(k)}$ vectors is 
\begin{equation}
\setlength{\abovedisplayskip}{3pt}
\setlength{\belowdisplayskip}{3pt}
    cycle\_vpu_{(k)} =  \lceil \frac{ S_{(k)} \times N_{(k)}}{m\times 16} \rceil
\end{equation}

\noindent For the $k^{th}$ layer, we assume the required cycles to finish  aggregation using the pipelined CirCore and VPU is: 
$$ cycle_{(k)}= max \left\{
\begin{array}{l}
cycle\_\mathit{fft}_{(k)}           \\
cycle\_mac_{(k)}          \\
cycle\_\mathit{ifft}_{(k)}          \\
cycle\_vpu_{(k)} 
\end{array}
\right.
$$
\noindent According to the profiling results in Table~\ref{tab:gnn_profile}, the aggregation time of GS-Pool
dominates the whole GNN computation. Therefore, the overall cycles can be estimated with the following formulation:
\begin{equation}
\setlength{\abovedisplayskip}{3pt}
\setlength{\belowdisplayskip}{3pt}
  cycle\_total  \approx  \sum_{k=1}^K cycle_{(k)} \times |\mathcal{V}|
\end{equation}

We assume the number of DSPs required to implement a single-channel FFT core is $\beta(n)$, and the systolic array requires $\gamma(l)$ DSPs for each PE. The SIMD-16 VPU with $m$ lanes consumes $m\times \eta$ DSPs. We then derive the following resource constraint: 
\begin{equation}
\setlength{\abovedisplayskip}{4pt}
\setlength{\belowdisplayskip}{4pt}
\label{DSP_constrain}
    \beta(n) \times (x+y) + r \times c \times \gamma(l) + m\times \eta \leq \mathit{\#DSPs}
\end{equation}

\noindent Given a GNN model and input graph, we can traversal search  all of the legal configurations and choose the optimal parameters with the minimal $cycle\_total$. With the small search space, the searching procedure only takes less than one minute on a desktop PC.  
\vspace{-0.2em}



\section{Evaluation}


\subsection{Experimental Setup}

\noindent$\bullet$ \textbf{Benchmarking Datasets and GNN Models:}~Table~\ref{tab:graph} lists the benchmarking graph datasets  including Reddit (RD), Cora (CR), Citeseer (CS), and Pumbed (PB). We use the four GNN models in Table ~\ref{tab:gnn} to evaluate our accelerator. For each model, we set $K=2$ and  set the dimention of the hidden-vectors  to 512. We adopt the random sampling strategy where the sample sizes $S_1=25$ and $S_2=10$.  

\begin{table}[t]
\centering
\caption{Graph Datasets}
\vspace{-1.4em}
\label{tab:graph}
\footnotesize
\setlength{\tabcolsep}{2.5mm}{
\renewcommand{\arraystretch}{0.99}{
\begin{center}
\begin{tabular}{|l|c|c|c|c|}
\hline
Graph       & \#Nodes & \#Edges    & \#Features & \#Labels \\ 
\hline
Cora (CR) & 2,708&  10,556 &   1,433 & 7 \\
Citeseer (CS)   & 3,327    & 4,732       & 3,703       & 6     \\
Pubmed (PB) & 19,717   & 44,338      & 500        & 3     \\
Reddit (RD) & 232,965 & 11,606,919 & 602        & 41    \\ 
\hline
\end{tabular}
\end{center}
}
}
\end{table}

\noindent$\bullet$\textbf{ Architectures for Comparison:} To demonstrate the efficiency of BlockGNN  and the proposed performance and resource model, we compare four different architectures: \textcircled{1}  BlockGNN-base: BlockGNN architecture with a set of  fixed hardware parameters for all the GNN tasks. \textcircled{2} BlockGNN-opt: BlockGNN architecture with the optimal hardware parameters for each  GNN task, according to  our performance and resource model. \textcircled{3} Intel Xeon(R) Gold 5220 CPU platform: The CPU server runs the uncompressed GNN models using the Tensorflow-based GraphSAGE framework. This platform equips 512GB DDR4 memory. 
\textcircled{4} HyGCN: The HyGCN architecture for comparison  uses a 6-lane SIMD-16 VPU and a systolic array ($4\times 32$) to compute aggregation and combination individually. Note that we  have reduced its scale to implement it on the same FPGA platform.
\vspace{-0.2em}
\subsection{Prototype Implementation}

\begin{table}[t]
\footnotesize
\centering
\caption{The Searched Optimal Parameters For GS-Pool}
\vspace{-1.5em}
\label{searched}
\setlength{\tabcolsep}{3.1mm}{

\begin{center}
\begin{tabular}{|c|cccccc|c|}

\hline
Dataset  & $x$ & $y$ & $r$ & $c$ & $l$ & $m$ &  min\_cycles\\ \hline
CR     &  18   & 7    &  6   &  4   &    1 & 1  & 24.9M    \\
CS    &    21 & 4     & 6    &  4   &    1 & 1  & 64.4M    \\
PB     &   14  &  15   &  4   &  4   &  1   &   1  &  95.4M  \\
RD    & 15    &    13 &    5 & 4    & 1    & 1  & 1240.3M     \\ \hline
\end{tabular}
\end{center}
}
\vspace{-1.2em}
\end{table}

To evaluate our design, we implement a prototype on  Xilinx ZC706 FPGA platform  using Vivado HLS (v2019.2).  We leverage the provided FFT IP (v9.1) to implement CirCore and realize multi-channel processing through paralleling multiple FFT/IFFT instances. The WB is set to $256$KB, which is large enough to store the compressed GNN model. We allocate  $512$KB for the NFB. 
To get  the remaining hardware parameters, we estimate the latency and DSP utilization coefficients  required by the performance and resource model.
When $n=128$ and using 32bit fixed-point,  we determine that  the FFT/IFFT cycles $\alpha(n)=484$. The number of DSPs  used to implement a single FFT/IFFT channel $\beta(n)$ is $18$. For each PE in the systolic array,  computing $l$ element-wise complex-number MAC per-cycle requires at least  $16\times l$ DSPs. Finally, each SIMD-16 lane consumes $64$ DSPs, thus we assume $\eta=64$. 

With these coefficients,  
we develop a tool to search for the optimal hardware parameters on the GNN models and datasets shown in Table \ref{tab:gnn} and Table \ref{tab:graph}, respectively.
\begin{figure*} [ht]
\setlength{\abovecaptionskip}{-0.1cm} 
\centering
    \includegraphics[width=1\linewidth]{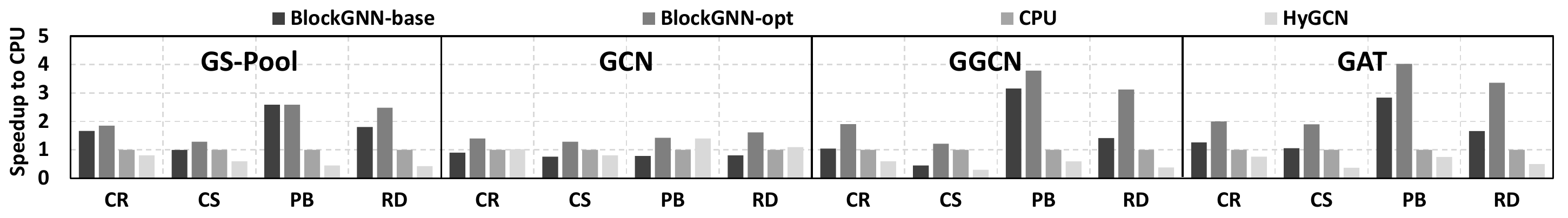} 
      \caption{Performance Comparisons.}
            \label{performance}
            \vspace{-1.2em}

\end{figure*}
The searched configuration for the GS-Pool model  is listed in Table~\ref{searched} as a representative example.
For different datasets, the  optimal FFT/IFFT channels $x,y$ and systolic array shape $r,c$ vary a lot. However, the parallelism of each PE ($l$) and VPU lanes ($m$) is always 1.  This indicates that the FFT/IFFT stages are the bottlenecks  for GS-Pool model, demanding more computational resources.  According to the optimal configuration, we also determine the parameters of BlockGNN-base, which has $16$ FFT/IFFT channels, $4\times 4$ systolic array, while $l$ and $m$ are set to 1.

After configuring the design, we export the RTL codes, which are synthesized and implemented using Vivado (v2019.2). The final working frequency is 100MHz. The FPGA resource utilization of BlockGNN-opt for GS-Pool model is listed in Table~\ref{resource}.   Note, all of the configurations make full use of the DSP resources,  indicating that it is appropriate  to use \#DSPs to constrain the searching procedure. 



\begin{table}[tbp]
\centering
\caption{The FPGA Resource Utilization For GS-Pool}
\vspace{-1.3em}
\label{resource}
\setlength{\tabcolsep}{3.35mm}{
\footnotesize
\begin{center}

\begin{tabular}{|c|c|c|c|c|}

\hline
Resource & BRAM\_18K & DSP48 & FF     & LUT    \\ \hline
Total     & 1090 & 900 & 437200 & 218600 \\ \hline
CR       &   39.3\%   &  99.8\%    &    27.7\%    &     34.6\%   \\
CS   &  41.8\%    &  99.8\%   &    35.3\%    &     44.8\%   \\
PB     &  42.2\%   &   93.6\% &  36.1\%      &   32.2\%     \\
RD       &   42.9\%   &  98.7\%  & 39.1\%      &    45.3\%    \\ \hline
\end{tabular}
\end{center}
}
\vspace{-1em}
\end{table}


\subsection{Performance Comparisons}
Figure~\ref{performance} depicts the performance comparisons of the four architectures on the benchmarking datasets.  Note that we ignore the neighbor sampling overhead and focus on the GNN inference time. The RD dataset exceeds the ZC706's DRAM capacity, so we partition it into two sub-graphs for evaluation.  The performance is measured by the normalized end-to-end execution time. Thanks to the block-circulant compression, on most of the tasks, both BlockGNN-base and BlockGNN-opt show considerable speedup. Compared to Xeon CPU and HyGCN,  BlockGNN-opt achieves on average $2.3\times$ and $4.2\times$ speedup respectively. On  G-GCN  and RD dataset, BlockGNN-opt achieves up to $8.3\times$ speedup against HyGCN.
Compared to BlockGNN-base, BlockGNN-opt usually has much better performance,  which demonstrates the effectiveness of our performance and resource model. The speedup on GCN is not as high as the other models, because the aggregation of GCN is not computation-intensive and the benefit of weight compression are not obvious. 

\subsection{Energy-efficiency Comparisons}
\begin{figure} [t]
\setlength{\abovecaptionskip}{-0.01cm} 
    \centering
    \includegraphics[width=0.98\linewidth]{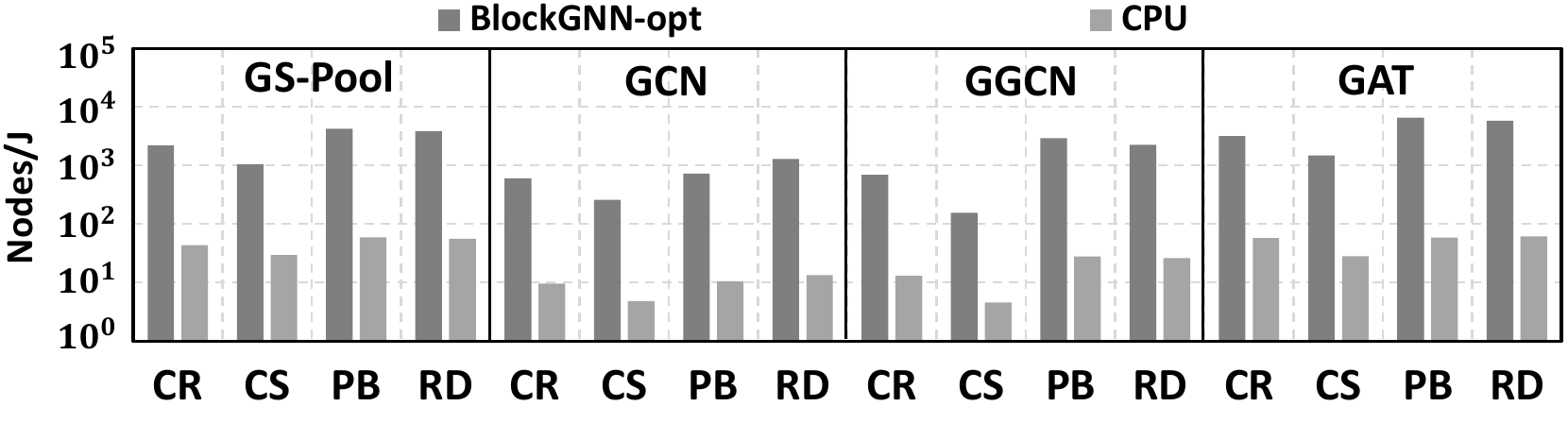} 
      \caption{Energy-efficiency Comparisons.}
                \label{fig:energy-eff}
    \vspace{-0.5em}
\end{figure}

 We measure the  power consumption of BlockGNN-opt and then compare it with the CPU baseline. The power of Xeon 5220 CPU is estimated as 125W, while the  power of BlockGNN-opt  is about 4.6W. We use the amount of total processed nodes and execution time to calculate \emph{Nodes-Per-Joule} (Nodes/J)  as the energy-efficiency metric.  As shown in Figure \ref{fig:energy-eff}, compared to the baseline CPU, BlockGNN-opt saves $33.9\times$ to $111.9\times$ energy, $68.9\times$ on average. Our design demonstrates great energy-efficiency, which is suitable for many power-sensitive edge-computing scenarios. 



\section{Discussion}

\noindent$\bullet$~\textbf{Use RFFT for Higher Speedup:}
We notice that there is still a gap between the implemented speedup (up to $8.3\times$) and the theoretical speedup (up to $18.3\times$). This is mainly because that the  FFT implementation using Xilinx IP can not get the ideal performance. However, 
for GNNs the inputs are always real values, so we don't have to use complex number as inputs. By using RFFT (Real-valued FFT) and IRFFT, the total computation can be greatly reduced. This observation has also been applied in previous work~\cite{CirCNN}.

\noindent$\bullet$~\textbf{Only Compress the Aggregators for Better Accuracy:}
In our experiments, we assume that both the weight matrices in combination and aggregation phases are compressed using block-circulant weight matrices, which will bring best compression ratio as well as simplifying hardware design. However, we also find that the accuracy will be much higher (less than 0.5\% accuracy drop) if we  only compress the aggregators. We leave it as another trade-off point.


\section{conclusion}

In this paper we propose BlockGNN, a software-hardware co-design approach to realize efficient GNN  acceleration. At the algorithm level, we  leverage block-circulant weight matrices  to reduce the computational complexity of various GNN models. At the hardware design level, we 
propose a novel BlockGNN  accelerator to  compute the compressed GNNs with low latency and high energy-efficiency. Moreover, to determine the optimal configurations for different tasks, we also introduce a performance and resource model. FPGA-based experiments demonstrate that compared to the baseline HyGCN architecture and Intel Xeon CPU, our design achieves up to $8.3\times$ speedup and $111.9\times$ energy reduction, respectively.
\section*{Acknowledgements}
This work is supported by National Key R\&D Program of China (2020AAA0105200), National Natural Science Foundation of China (Grant No.61832020 \& 62032001) and China National Postdoctoral Program for Innovative Talents (No. BX20200001). We also thank Dimin Niu and Hongzhong Zheng for their help.

\bibliographystyle{IEEEtran}
\bibliography{reference}

\end{document}